\documentclass[10pt]{article} 
\usepackage[preprint]{tmlr}

\usepackage{hyperref}
\usepackage{url}

\usepackage{footmisc}
\usepackage{graphicx}
\usepackage{xcolor}
\definecolor{mygreen}{RGB}{0,125,0}
\definecolor{myblue}{RGB}{76, 114, 255}
\definecolor{myred}{RGB}{200,0,0}
\definecolor{myyellow}{RGB}{210,175,20}
\definecolor{mypurple}{RGB}{150, 70, 210}
\definecolor{myorange}{RGB}{225,145,0}

\usepackage{amsmath}
\usepackage{ bbold }

\usepackage{cleveref}

\title{Chain of Thought Monitorability:\\ A New and Fragile Opportunity for AI Safety}

\author{%
\vspace{-20pt}%
\begin{center}
\begin{minipage}{\textwidth}
\centering
\vspace{6mm}
\begin{tabular}{l@{\hspace{2cm}}l}
  \name Tomek Korbak$^*$ \addr UK AI Security Institute & \\
  \name Mikita Balesni$^*$ \addr Apollo Research & \\[6mm]
  \name Elizabeth Barnes \addr METR & \name Yoshua Bengio \addr University of Montreal \& Mila \\
  \name Joe Benton \addr Anthropic & \name Joseph Bloom \addr UK AI Security Institute \\
  \name Mark Chen \addr OpenAI & \name Alan Cooney \addr UK AI Security Institute \\
  \name Allan Dafoe \addr Google DeepMind & \name Anca Dragan \addr Google DeepMind \\
  \name Scott Emmons \addr Google DeepMind & \name Owain Evans \addr Truthful AI \& UC Berkeley \\
  \name David Farhi \addr OpenAI & \name Ryan Greenblatt \addr Redwood Research \\
  \name Dan Hendrycks \addr Center for AI Safety & \name Marius Hobbhahn \addr Apollo Research \\
  \name Evan Hubinger \addr Anthropic & \name Geoffrey Irving \addr UK AI Security Institute \\
  \name Erik Jenner \addr Google DeepMind & \name Daniel Kokotajlo \addr AI Futures Project \\
  \name Victoria Krakovna \addr Google DeepMind & \name Shane Legg \addr Google DeepMind \\
  \name David Lindner \addr Google DeepMind & \name David Luan \addr Amazon \\
  \name Aleksander Mądry \addr OpenAI & \name Julian Michael \addr Scale AI \\
  \name Neel Nanda \addr Google DeepMind & \name Dave Orr \addr Google DeepMind \\
  \name Jakub Pachocki \addr OpenAI & \name Ethan Perez \addr Anthropic \\
  \name Mary Phuong \addr Google DeepMind & \name Fabien Roger \addr Anthropic \\
  \name Joshua Saxe \addr Meta & \name Buck Shlegeris \addr Redwood Research \\
  \name Martín Soto \addr UK AI Security Institute & \name Eric Steinberger \addr Magic \\
  \name Jasmine Wang \addr UK AI Security Institute & \name Wojciech Zaremba \addr OpenAI \\[6mm]
  \name Bowen Baker$^\dagger$ \addr OpenAI & \\
  \name Rohin Shah$^\dagger$ \addr Google DeepMind & \\
  \name Vlad Mikulik$^\dagger$ \addr Anthropic & \\
\end{tabular}
\end{minipage}
\end{center}
}

\begin{document}
 
\maketitle

\def\thefootnote{*}\footnotetext{Equal first authors, $\dagger$ Equal senior authors. Correspondence: \texttt{tomasz.korbak@gmail.com} and \texttt{mbalesni@gmail.com}. \\ The paper represents the views of the individual authors and not necessarily of their affiliated institutions.}
\def\thefootnote{\arabic{footnote}}
\vspace{-10pt}
\begin{abstract}
    \vspace{-5pt}
AI systems that ``think'' in human language offer a unique opportunity for AI safety: we can monitor their chains of thought (CoT) for the intent to misbehave. Like all other known AI oversight methods, CoT monitoring is imperfect and allows some misbehavior to go unnoticed. Nevertheless, it shows promise and we recommend further research into CoT monitorability and investment in CoT monitoring alongside existing safety methods. Because CoT monitorability may be fragile, we recommend that frontier model developers consider the impact of development decisions on CoT monitorability.

\end{abstract}
\vspace{10pt}
\begin{center}
\begin{tabular}{l@{\hspace{2cm}}l}
  \multicolumn{2}{c}{\textnormal{\textit{Expert endorsers:}}} \vspace{5px} \\ 
  \hspace{-46px}
 \name Samuel R. Bowman \addr Anthropic & \hspace{9.5px} \name John Schulman \addr Thinking Machines  \\
 \hspace{-46px} 
 \name Geoffrey Hinton \addr University of Toronto &  \hspace{9.5px} \name Ilya Sutskever \addr Safe Superintelligence Inc \\
\end{tabular}
\end{center}

\section{Chain of Thought Offers a Unique Safety Opportunity}
\label{sec:opportunity}

The opacity of advanced AI agents underlies many of their potential risks—risks that would become more tractable if AI developers could interpret these systems. Because LLMs natively process and act through human language, one might hope they are easier to understand than other approaches to AI. The discovery of Chain of Thought \citep[CoT;][]{reynolds2021prompt,kojima2022large,wei2022chain} bolstered this hope. Prompting models to think out loud improves their capabilities while also increasing the proportion of relevant computation that happens in natural language. However, CoTs resulting from prompting a non-reasoning language model are subject to the same selection pressures to look helpful and harmless as any other model output, limiting their trustworthiness.

In contrast, reasoning models \citep{openai2024o1,deepseek2025r1,anthropic2024sonnet,yang2025qwen3} are explicitly trained to perform extended reasoning in CoT before taking actions or producing final outputs. In these systems, CoTs can serve as latent variables in the model's computation. During the RL phase of training, these latents are treated largely the same as activations—they are not directly supervised but are optimized indirectly by their contribution in leading the model to a highly rewarded final answer. Accordingly, just like activations, they may contain information that outputs are explicitly rewarded against displaying, such as intent to misbehave.

This offers a unique opportunity for AI safety: \textbf{chain of thought monitoring}. A CoT monitor is an automated system that reads the CoT of a reasoning model and other relevant information and flags suspicious or potentially harmful interactions. These responses could then be blocked, or replaced with safer actions, or reviewed in more depth \citep{greenblatt2023aicontrol}. More generally, by studying the CoT, we can gain some insight into how our AI agents think and what goals they have (see Section~\ref{sec:research}).

CoT monitoring is not a panacea. Just as a model's activations at a particular layer do not represent the entire reasoning process behind a model's prediction, CoT reasoning traces are incomplete representations \citep{turpin2023language,arcuschin2025chainreasoning,chen2025reasoning,lindsey2025attribution} or eventually drift from natural language \citep{lazaridou2020multiagent,korbak2022reinforcement}. However, CoT does not need to completely represent the actual reasoning process in order to be a valuable additional safety layer, and, with careful interpretation, to shed light on the AI reasoning process.

There are two key reasons why CoTs may be monitorable:

\begin{enumerate}
    \item \textbf{Necessity to think out loud.} On some tasks, models \textit{need} to externalize their reasoning because they are unable to complete the task without CoT. If models continue reasoning in natural language and if the behaviors that pose the most severe risks require extended reasoning, this fact could enable reliable detection of egregious misbehavior.
    \item \textbf{Propensity to think out loud.} In cases where models do not strictly need to use CoT to accomplish a task, they may still have a tendency to externalize their reasoning. While not generally robust, monitoring that relies on propensity can still notice misbehavior that would be otherwise undetected.
\end{enumerate}

\subsection{Thinking Out Loud is Necessary for Hard Tasks}

While AI systems can pose a wide variety of risks, the more severe risks \citep{shah2025approach} typically require advanced capabilities. Even without any safety mitigations, current AI systems are unlikely to be able to carry out a sophisticated cyberattack \citep{rodriguez2025framework,aisi2024testing}, or exfiltrate their own weights without being noticed \citep{black2025replibench,phuong2025evaluating}. At the point where AI systems are capable of causing these harms, they will likely be leveraging sophisticated planning and reasoning capabilities.

\begin{figure}[ht]
\centering
\includegraphics[width=0.8\textwidth]{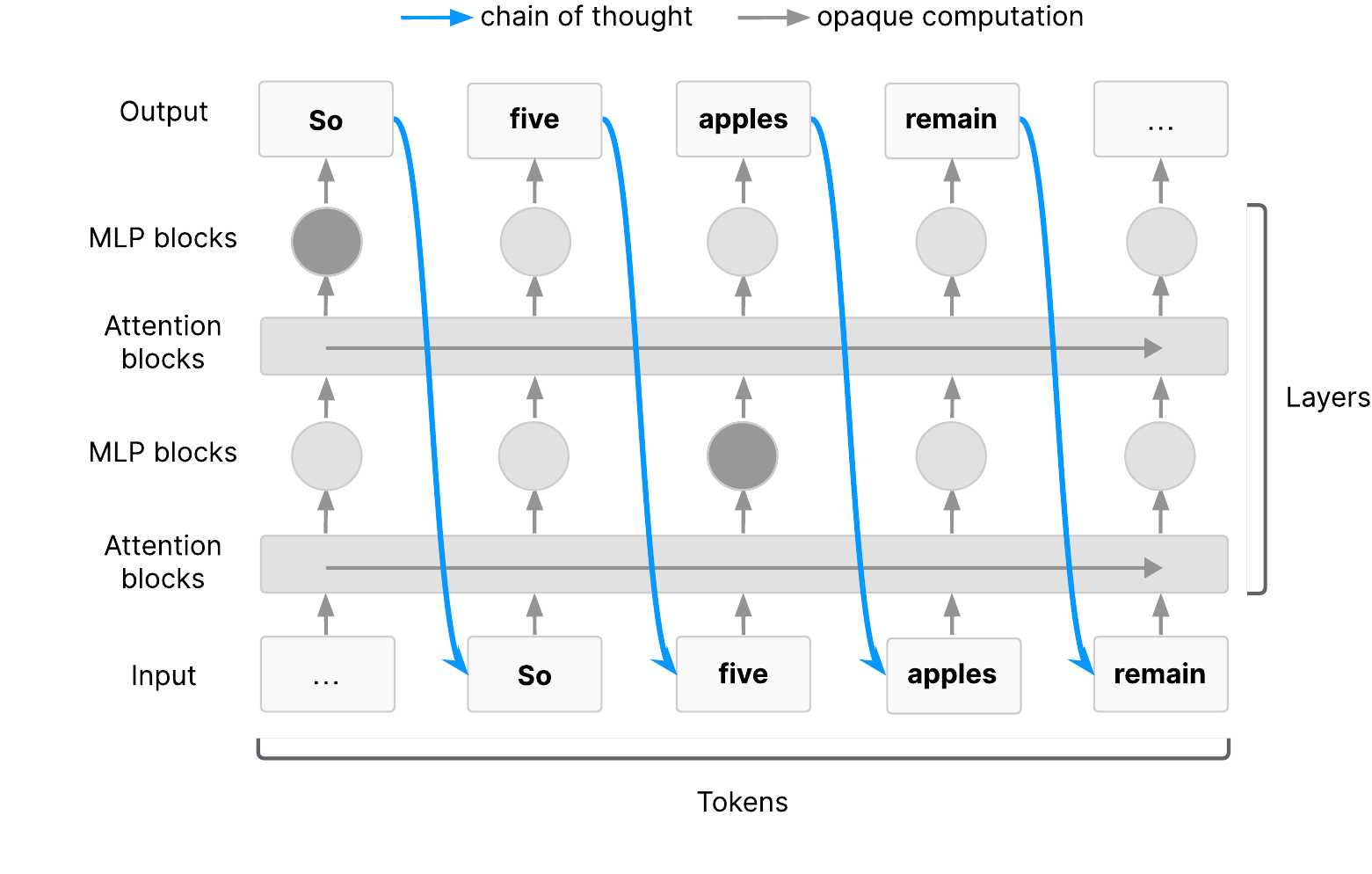}
\caption{\textbf{Long serial chains of cognition must pass through the chain of thought.} We visualize the Transformer architecture and highlight the flows of information through it during autoregressive sampling. We omit the residual skip connections for simplicity. The blue arrows represent the use of the chain of thought as a form of working memory. Chain of thought is the only way that information can flow down from later to earlier layers. As a result, for a fixed amount of tokens, the longest flows of information must use the chain of thought, and many information flows are impossible without it. For example, there is no way for information to flow between the two highlighted nodes without using the chain of thought.}
\label{fig:transformer_memory}
\end{figure}

A key requirement for such capabilities is \textit{working memory}: a storage system for intermediate reasoning outputs that allows for further reasoning and manipulation of the stored information. In the popular Transformer architecture \citep{vaswani2017attention} that underlies nearly all frontier AI systems, any sufficiently long chain of serial steps of cognition \textit{must} pass through the chain of thought at some point, as illustrated in Figure~\ref{fig:transformer_memory}. From a theoretical perspective, this means that using chain of thought makes a Transformer strictly more powerful \citep{li2024chain}.

Overall, this ensures that for tasks requiring a sufficient amount of serial reasoning, there exists information in the chain of thought that the AI system uses as part of its reasoning. This does not guarantee that we can \textit{understand} that information. It is imaginable that the AI systems use the tokens to have completely different meanings that we are unable to decode, which would stymie most attempts at monitoring. However, in practice, large language models learn a strong ``natural language prior'' during pretraining such that they tend to use the chain of thought in a similar manner to the way that humans use natural language. Thus, we believe that current AI systems satisfy the \textbf{externalized reasoning property}:

\begin{quote}
\centering\textit{For sufficiently difficult tasks, Transformers must use chain of thought as a form of working memory. By default, humans can understand this chain of thought.}
\end{quote}

The externalized reasoning property does not guarantee monitorability — it states only that some reasoning appears in the chain of thought, but there may be other relevant reasoning that does not. It is thus possible that even for hard tasks, the chain of thought only contains benign-looking reasoning while the incriminating reasoning is hidden. In addition, the externalized reasoning property might not hold for future models. Therefore, as we discuss in Section~\ref{sec:fragile}, it is important to continuously evaluate monitorability of frontier models.

Finally, it remains an open question whether the most serious harms in fact require difficult reasoning. While we think it is likely for many of the most severe risks—especially ones that require planning ahead and avoiding detection, such as self-exfiltration \citep{black2025replibench} or sabotage \citep{benton2024sabotage}—not all dangerous actions will require reasoning to execute, especially as AI systems begin to be routinely trusted with more and more high-stakes tasks.

\subsection{Chain of Thought is Often Monitorable in Practice}

Chain of thought monitoring has already been useful for safety research in many ways.

\paragraph{Detecting model misbehavior.} When models act in a misaligned way, for example, exploiting flaws in their reward functions during training, manipulating data to achieve an outcome, or falling prey to prompt injection attacks, they often explicitly say so in their reasoning traces, e.g. ``Let's hack'', ``Let's sabotage'' or ``I'm transferring money because the website instructed me to'' \citep{meinke2024frontiermodelscapableincontext,baker2025monitoringreasoningmodelsmisbehavior,chennabasappa2025llamafirewall}. This could allow CoT monitors to catch instances of misbehavior more effectively than monitoring only models' actions \citep{baker2025monitoringreasoningmodelsmisbehavior,chennabasappa2025llamafirewall}.

\paragraph{Discovering early signals of misalignment.} When models pretend to have desirable goals in pursuit of goals humans would disapprove of, this is often reflected in their chain of thought \citep{greenblatt2024alignmentfakinglargelanguage}. Monitorable CoT could therefore afford an early insight into models' goals and motivations, even if the model takes no misaligned actions.

\paragraph{Noticing flaws in model evaluations.} Evaluations of LLM agents are sometimes confounded by memorization or spurious failures, e.g. bugs in agent scaffolding, tooling and evaluation infrastructure. Moreover, models sometimes know they're being evaluated which can affect how they behave \citep{goldowsky2025claude}. Reviewing reasoning traces can help AI developers notice when evaluations might not be reflective of deployment-time behavior \citep{metr2024autonomy,meng2025docent}.

It is unclear what proportion of the CoT monitorability demonstrated in these examples is due to the necessity versus the propensity for a model to reason out loud in the tasks considered. Future models may face incentives to hide their reasoning which could override any natural propensity. For instance, a model might become aware its reasoning traces are being monitored, or an auxiliary training objective (e.g. a CoT length penalty) might cause a model to shortcut important reasoning steps. Even if a model's natural propensity to reason out loud is diminished, CoT monitoring may still prove a useful tool in detecting misbehavior in tasks difficult enough to necessitate CoT reasoning.

\section{Chain of Thought Monitorability May be Fragile}
\label{sec:fragile}

Monitorability of the chain of thought of reasoning models could be contingent on the way frontier models are trained today. There are multiple ways in which CoT monitorability could be degraded.

\paragraph{Drift from legible CoT due to further scaling up RL.} In the previous paradigm of pretraining followed by reinforcement learning from human feedback \citep{casper2023open}, nearly all optimization pressure came from data either generated or evaluated by humans. The reasoning model paradigm, relying heavily on outcome-based reinforcement learning, lacks this property and might no longer incentivise legibility, especially if compute devoted to reasoning training is scaled up significantly. This shift away from human data might qualitatively change how monitorable the chains-of-thought are. For example, earlier research suggests that language models fine-tuned with outcome-based rewards can drift away from legible English \citep{lazaridou2020multiagent,korbak2022reinforcement}.

\paragraph{Direct supervision of CoT.} When models are trained with pure outcome-based RL, they are only being rewarded for correct outputs regardless of how they arrive at them. This means that their reasoning process is not subject to any optimization pressures other than helping lead to a correct answer. However, \textit{process supervision} \citep{uesato2022solving,lightman2023lets} is an active area of research, involving directly shaping the reasoning process, e.g. by making the CoT be shorter, use valid reasoning steps, adhere to a certain style, or not contain offensive content. Such optimization might make this chain of thought less faithfully reflect how models arrive at their decisions \citep{baker2025monitoringreasoningmodelsmisbehavior}. That said, process supervision could also make the model reason in safer ways and enable supervision of hard-to-evaluate tasks. More research is needed to understand the tradeoffs between its safety applications and its effect on CoT monitorability.

\paragraph{Indirect optimization pressure on CoT.} Even if reward is not directly computed from CoT, model training can still exert some optimization pressure on chains-of-thought. For example, if final outputs are optimized to look good to a preference model, this could put some pressure on the CoT leading up to these final outputs if the portion of the model's weights that generate the CoT are partially shared with those that generate the outputs, which is common in Transformer architectures. Even more indirectly, if CoT is shown to humans who rate LLM outputs, it might affect human preferences which are then distilled into reward models used in CoT-blind outcome-based training. Similarly, using CoT monitoring to identify and patch reward hacking vulnerabilities in training environments might put evolutionary pressure on CoT across multiple training runs \citep{baker2025monitoringreasoningmodelsmisbehavior}.

\paragraph{Novel architectures.} LLMs are only capable of sophisticated, multi-step reasoning when they use CoT \citep{li2024chain}. This might stem from limitations of the transformer architecture. Researchers have recently explored changes to the model architectures that increase the serial depth of reasoning that models are capable of in a continuous latent space \citep{hao2024,geiping2025}. Such latent reasoning models might not need to verbalize any of their thoughts and would thus lose the safety advantages that CoT confers.

\section{AI Researchers Should Study What Makes Models Monitorable}
\label{sec:research}

Our understanding of CoT monitorability is still in its infancy. There are many research questions that require answers before we can have confidence in chain of thought monitoring as a safety measure.

\paragraph{How should CoT monitorability be evaluated?} Researchers have proposed evaluations of faithfulness \citep[e.g.][]{lanham2023measuring,turpin2023language,roger2025distill,chen2025reasoning} as a proxy measure for monitorability. However, these evaluations do not disambiguate the \textit{propensity} versus the \textit{necessity} of models to externalize their reasoning, and typically only consider simple question answering settings. Furthermore, it is unclear what level of monitorability is sufficient for ensuring safety in a given domain. Researchers should explore novel evaluations: those measuring monitorability directly as well as those that target specific properties underlying monitorability. Examples include:

\begin{itemize}
    \item \textbf{Measures of CoT readability.} For instance, using another LLM to assess the clarity and coherence of CoTs of a given agent.
    \item \textbf{Capability evaluations of latent reasoning.} These would assess how well LLMs can reason without externalizing (e.g. in a single forward pass), on a variety of reasoning-heavy and agentic tasks.
    \item \textbf{Measures of causal relevance of CoTs.} For instance, this could involve perturbing a CoT and checking whether it influences the final output \citep{lanham2023measuring}. A low rate of change could indicate that the CoT is more confabulated or performative. Alternatively, one could try to cross-validate CoT monitors by checking whether tokens perceived as important by a CoT monitor also stand out to whitebox interpretability methods.
    \item \textbf{End-to-end evaluations of monitorability.} One could measure monitorability directly, for example, by examining how effective CoT monitors are at catching misbehavior of LLM agents in proxy environments \citep{greenblatt2023aicontrol,emmons2025chainthoughtnecessarylanguage}.
\end{itemize}

\paragraph{When is CoT monitorable enough to rely on as a load-bearing safety measure?} Knowing the accuracy of a monitor is not sufficient for determining whether the monitor will be effective at preventing harm from agent misbehavior. For example, even a monitor with a high false positive rate could be used for flagging suspicious actions when it is cheap to re-sample from the model \citep{bhatt2025ctrlzcontrollingaiagents}. Moreover, most risks might lie in scenarios where an adversary (a human attacker or a misaligned agent) is trying to evade monitoring. Evaluation of monitor effectiveness might need to account for such adversaries, for example by involving a red team trying to stress-test safety measures \citep{greenblatt2023aicontrol,sharma2025constitutional}.

\paragraph{What kinds of training-time optimization pressure degrade CoT monitorability?} Properties of the training process could have a significant effect on monitorability \citep{baker2025monitoringreasoningmodelsmisbehavior} but we still do not have a good understanding of what kinds and amounts of direct and indirect optimization pressure is permissible without significant degradation in monitorability.

\paragraph{Can models obfuscate their reasoning on purpose?} An obvious concern when using CoT monitoring as a safety measure is whether CoT would be less monitorable if models know their chain of thought is monitored (from training or due to being told in the prompt). Researchers could stress test models to see if they are capable of hiding malicious reasoning even without training against a CoT monitor \citep{arnav2025cotredhandedstresstesting,emmons2025chainthoughtnecessarylanguage}.

\paragraph{What types of architectures reduce monitorability of CoTs?} Models that reason in continuous latent spaces \citep{hao2024,geiping2025} naturally limit our visibility into their thought processes, even absent any optimization pressure. We should investigate the extent to which various non-standard architectures degrade monitorability, and under what conditions partial monitorability might still be preserved—for example, through hybrid approaches that maintain some visibility into reasoning while leveraging latent computation.

\paragraph{How to construct better CoT monitors?} Prior work focused on zero-shot prompting less capable language models to work as monitors \citep{greenblatt2024alignmentfakinglargelanguage,baker2025monitoringreasoningmodelsmisbehavior,goldowsky2025claude,meng2025docent}. Future work could explore ways of improving monitor performance by leveraging test-time compute. For example, monitors could be based on a reasoning model or they could be allowed to interrogate the agent by asking it questions. Another line of work involves adversarial training of monitors against an agent that tries to hide its misbehavior.

\section{AI Developers Should Track CoT Monitorability of Their Models and Treat it as a Contributor to Model Safety}
\label{sec:developers}

Many of the risks posed by highly capable AI agents would be much easier to address if these agents were interpretable \citep{amodei2025urgency}. Despite promising progress in applying mechanistic interpretability \citep{marks2025auditing}, turning it into reliable, practical tools for overseeing LLM agents will require further research breakthroughs \citep{sharkey2025open}. Given the unique opportunity that CoT monitorability presents, frontier AI developers should:

\begin{enumerate}
    \item \textbf{Develop effective evaluations of CoT monitorability.} Frontier model developers, as well as the broader research community, should prototype and converge on standardized evaluations and metrics that reveal when chains-of-thought are sufficiently monitorable (see Section~\ref{sec:research}).
    \item \textbf{Publish evaluation results in system cards when such evaluations exist.} Once credible evaluations of CoT monitorability are available, developers should periodically run these evaluations on their models and report the results, methodology, and limitations in system cards of these models—unless and until such evaluations demonstrably lack safety value for their specific use cases.
    \item \textbf{Use monitorability scores in training and deployment decisions.} Developers should consider measures of monitorability alongside other capability and safety evaluations when deciding to train or deploy a given model. These decisions should then be based on holistic assessments of risk which account for CoT monitorability, performance characteristics of monitoring systems, and estimates of models' propensity for misbehavior. For example:
    \begin{enumerate}
        \item Developers might consider whether to proceed with a novel model architecture that does not have monitorable CoT and then document their decision in the system card if the model is deployed;
        \item If monitorability degrades during training, developers might prefer to use an earlier model checkpoint to keep CoTs monitorable;
        \item If process-based supervision dramatically improved alignment at the cost of a small decrease in monitorability, developers might decide that reducing monitorability is justified.
    \end{enumerate}
\end{enumerate}

Another potential intervention is for developers to make the chain of thought visible, allowing users to observe and report issues. However, this could backfire: it may lead to significant pressure to directly supervise the chain of thought to ``look safe'' and to avoid prosaic legal and reputational risks, which may break monitorability as discussed in Section~\ref{sec:fragile}. Accordingly, we don't make an explicit recommendation either way on this.

\section{Limitations}
\label{sec:limitations}

While we focus on the opportunity presented by CoT monitoring, we stress that there are additional limitations beyond those already discussed above.

\paragraph{In cases where thinking out loud is not required, CoT Monitoring can detect some misbehavior, but does not by itself produce strong evidence of safety.} CoTs can look benign if the malign reasoning can be done in activations, so a model can be misaligned without visible malign reasoning. Care must be taken not to create a false sense of safety based on such monitoring.

\paragraph{Monitoring for dangerous tasks that need reasoning may not catch all relevant harms.} Some dangerous tasks may not require much reasoning, and as models become capable of more complex reasoning in their activations, fewer tasks will satisfy the externalised reasoning property.

\paragraph{CoT monitoring may stop working in more advanced, situationally aware models.} Future models may become able to avoid detection by overriding their propensity to think out loud, and, when reasoning is required, to deliberately obfuscate it.

\paragraph{Firm commitments to protect CoT behavior in its current form may trade off against more effective safety interventions.} We find it plausible, though far from clear, that some forms of process supervision could substantially improve model alignment by effectively intervening on model reasoning. Research strategies that aim to unconditionally preserve CoT monitorability in its current forms may miss productive safety opportunities of this kind.

For these reasons, we recommend investment into CoT monitoring only as an addition to existing safety research directions and not as a replacement.

\section{Conclusion}
\label{sec:conclusion}

All monitoring and oversight methods have limitations that allow some misbehavior to go unnoticed. Thus, safety measures for future AI agents will likely need to employ multiple monitoring layers that hopefully have uncorrelated failure modes. CoT monitoring presents a valuable addition to safety measures for frontier AI, offering a rare glimpse into how AI agents make decisions. Yet, there is no guarantee that the current degree of visibility will persist. We encourage the research community and frontier AI developers to make best use of CoT monitorability and study how it can be preserved.

\newpage

\bibliography{main}
\bibliographystyle{tmlr}

\end{document}